\documentclass{article}
\usepackage[utf8]{inputenc}

\usepackage{hyperref}

\usepackage[accepted]{pkg/icml2020}
\usepackage{times}
\usepackage{epsfig}
\usepackage{graphicx}
\usepackage{amsmath}
\usepackage{amssymb}
\usepackage{booktabs}
\usepackage{framed}
\usepackage{multirow}

\newcommand{\esp}{\mathbb{E}}
\newcommand{\prob}{\mathbb{P}}
\newcommand{\hyp}{\mathcal{H}}

\begin{document}

\definecolor{darkgreen}{RGB}{0, 140, 0}
\newcommand{\rv}[1]{{\color{darkgreen}[\textbf{Rv}:#1]}}
\newcommand{\alex}[1]{{\color{red}[\textbf{Alex}:#1]}}
\newcommand{\matthijs}[1]{{\color{blue}[\textbf{Matthijs}:#1]}}
\newcommand{\cordelia}[1]{{\color{magenta}[\textbf{Cordelia}:#1]}}

\icmltitlerunning{Radioactive data: tracing through training}

\twocolumn[
\icmltitle{Radioactive data: tracing through training}

\begin{icmlauthorlist}

\icmlauthor{Alexandre Sablayrolles}{fb,inria}
\icmlauthor{Matthijs Douze}{fb}
\icmlauthor{Cordelia Schmid}{inria}
\icmlauthor{Herv\'e J\'egou}{fb}

\end{icmlauthorlist}

\icmlaffiliation{fb}{Facebook AI Research}
\icmlaffiliation{inria}{INRIA}

\icmlcorrespondingauthor{Alexandre Sablayrolles}{alexandre.sablayrolles@gmail.com}

\icmlkeywords{Machine Learning, ICML}

\vskip 0.3in
]

\printAffiliationsAndNotice %

\begin{abstract} 
We want to detect whether a particular image dataset has been used to train a model.
We propose a new technique, \emph{radioactive data}, that makes imperceptible changes to this dataset such that any model trained on it will bear an identifiable mark. 
The mark is robust to strong variations such as different architectures or optimization methods.  
Given a trained model, our technique detects the use of radioactive data and provides a level of confidence ($p$-value).

Our experiments on large-scale benchmarks (Imagenet), using standard architectures (Resnet-18, VGG-16, Densenet-121) and training procedures, show that we can detect usage of radioactive data with high confidence ($p<10^{-4}$) even when only $1\%$ of the data used to trained our model is radioactive. 
Our method is robust to data augmentation and the stochasticity of deep network optimization. 
As a result, it offers a much higher signal-to-noise ratio than data poisoning and backdoor methods.
\end{abstract}

\section{Introduction}

The availability of large-scale public datasets has accelerated the development of machine learning. 
The Imagenet collection~\cite{deng2009imagenet} and challenge~\cite{russakovsky2015imagenet} contributed to the success of the deep learning architectures~\cite{krizhevsky2012imagenet}. 
The annotation of precise instance segmentation on the large-scale COCO dataset \cite{lin2014microsoft} enabled large improvements of object detectors and instance segmentation models \cite{he2017mask}.
Even in weakly-supervised~\cite{joulin2016learning,mahajan18eccv} and unsupervised learning \cite{caron19unsupervised} where annotations are scarcer, state-of-the-art results are obtained on large-scale datasets collected from the Web \cite{thomee2015yfcc100m}.

Machine learning and deep learning models are trained to solve specific tasks (e.g. classification, segmentation), but as a side-effect reproduce the bias in the datasets \cite{torralba2011unbiased}.
Such a bias is a weak signal that a particular dataset has been used to solve a task. 
Our objective in this paper is to enable the traceability for datasets. By introducing a specific mark in a dataset, we want to provide a strong signal that a dataset has been used to train a model. 
\begin{figure}[t]
\includegraphics[width=\linewidth]{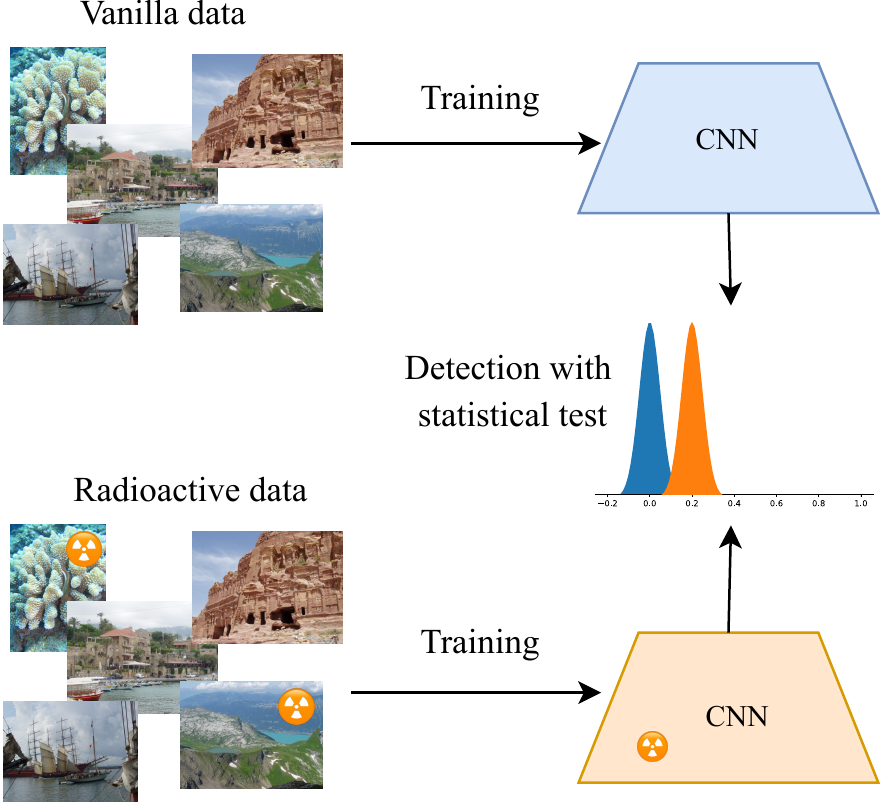}
\vspace{-15pt}
\caption{
Illustration of our approach: we want to determine through a  statistical test ($p$-value) whether a network has seen a marked dataset or not. 
The distribution (shown on the histograms) of a statistic on the network weights is clearly separated between the vanilla and radioactive CNNs. 
Our method works in the cases of both white-box and black-box access to the network.
\label{fig:splatch}
}
\vspace{-10pt}
\end{figure}
We thus slightly change the dataset, effectively substituting the data for similar-looking marked data (\emph{isotopes}).

Let us assume that this data, as well as other collected data, is used to train a convolutional neural network (convnet).
After training, the model is inspected to assess the use of radioactive data. 
The convnet is accessed either (1) explicitly when the model and corresponding weights are available (white-box setting), or (2) implicitly if only the decision scores are accessible (black-box setting).
From that information, we answer the question of whether any radioactive data has been used to train the model, or if only vanilla data was used.  
We want to provide a statistical guarantee with the answer, in the form of a $p$-value. 

\emph{Passive} techniques such as those employed to measure dataset bias~\cite{torralba2011unbiased} or to do membership inference~\cite{sablayrolles19whitebox,shokri17membershipinference} cannot provide sufficient empirical or statistical guarantees. 
More importantly, their measurement is relatively weak and therefore cannot be considered as an evidence: they are likely to confuse datasets having the same underlying statistics. %
In contrast, we target a $p$-value much below $0.1\%$, meaning there is a very low probability that the results we observe are obtained by chance.

Therefore, we  focus on \emph{active} techniques, where we apply visually imperceptible changes to the images. 
We consider the following three criteria: 
(1) The change should be tiny, as measured by an image quality metric like PSNR (Peak Signal to Noise Ratio);
(2) The technique should be reasonably neutral with respect to the end-task, i.e., the accuracy of the model trained with the marked dataset should not be significantly modified; 
(3) The method should not be detectable by a visual analysis of failure cases and should be immune to a re-annotation of the dataset.
This disqualifies techniques that employ incorrect labels as a mark, which are easy to detect by a simple analysis of the failure cases. 
Similarly the ``backdoor'' techniques are easy to identify and circumvent with outlier detection~\cite{tran18spectral}.

At this point, one may draw the analogy between this problem and watermarking~\cite{cox2002digital}, whose goal is to imprint a mark into an image such that it can be re-identified with high probability. 
We point out that traditional image-based watermarking is ineffective in our context: the learning procedure ignores the watermarks if they are not useful to guide the classification decision \cite{tishby2000information}. 
Therefore regular watermarking leaves no exploitable trace after training. 
We need to force the network to keep the mark through the learning process, whatever the learning procedure or architecture. 

To that goal, we propose \textit{radioactive data}. As illustrated in Figure~\ref{fig:splatch} and  %
similarly to radioactive markers in medical applications, we introduce marks (data \textit{isotopes}) that remain through the learning process and that are detectable with high confidence in a neural network. 
Our idea is to craft a \emph{class-specific} additive mark in the latent space before the  classification layer. 
This mark is propagated back to the pixels with a marking (pretrained) network. 

This behaviour is confirmed by an analysis of the latent space before classification. 
It shows that the network devotes a small part of its capacity to keep track of our ``radioactive tracers''. %

Our experiments on Imagenet confirm that our radioactive marking technique is effective: with almost invisible changes to the images ($\text{PSNR}=42~\text{dB}$), and when marking only a fraction of the images ($q=1\%$), we are able to detect the use of our radioactive images with very strong confidence. 
Note that our radioactive marks, while visually imperceptible, might be detected by a statistical analysis of the latent space of the network. 
Our aim in this paper is to provide a proof of concept that marking data is possible with statistical guarantees, and the analysis of defense mechanisms lies outside the scope of this paper. 
The deep learning community has developed a variety of defense mechanisms against ``adversarial attacks": these techniques prevent test-time tampering, but are not designed to prevent training-time attacks on neural networks. 

Our conclusions are supported in various settings: we consider both the black-box and white-box settings; we change the tested architecture such that it differs from the one employed to insert the mark. 
We also depart from the common restrictions of many data-poisoning works \cite{shafahi18poisonfrogs,biggio12poisoning}, where only the logistic layer is retrained, and which consider small datasets (CIFAR) and/or limited data augmentation. 
We verify that the radioactive mark holds when the network is trained from scratch on a radioactive Imagenet dataset with standard random data augmentations.
As an example, for a ResNet-18 trained from scratch, we achieve a $p$-value of $10^{-4}$ when only $1\%$ of the training data is radioactive.
The accuracy of the network is not noticeably changed ($\pm 0.1\%$).

The paper is organized as follows. Section~\ref{sec:related} reviews the related literature. We discuss related works in watermarking, and explain how the problem that we tackle is related to and differs from data poisoning. 
In Section~\ref{sec:method}, after introducing a few mathematical notions, we describe how we add markers, and discuss the detection methods in both the white-box and black-box settings. Section~\ref{sec:analysis} provides an analysis of the latent space learned with our procedure and compares it to the original one. We present qualitative and quantitative results in different settings in the experimental section~\ref{sec:experiments}. We conclude the paper in Section~\ref{sec:conclusion}.

\section{Related work}
\label{sec:related}

\paragraph{Watermarking}
is a way of tracking media content by adding a mark to it. 
In its simplest form, a watermark is an addition in the pixel space of an image, that is not visually perceptible.
Zero-bit watermarking techniques \cite{cayre2005watermarking} modify the pixels of an image so that its Fourier transform lies in the cone generated by an arbitrary random direction, the ``carrier".
When the same image or a slightly perturbed version of it are encountered, the presence of the watermark is assessed by verifying whether the Fourier representation lies in the cone generated by the carrier. 
Zero-bit watermarking detects whether an image is marked or not, but in general watermarking also considers the case where the marks carry a number of bits of information~\cite{cox2002digital}. 

Traditional watermarking is notoriously not robust to geometrical attacks \cite{vukotic2018deep}. 
In contrast, the latent space associated with deep networks is almost invariant to such transformations, due to the train-time data augmentations. 
This observation has motivated several authors to employ convnets to watermark images ~\cite{vukotic2018deep,zhu2018hidden} by inserting marks in this latent space. HiDDeN~\cite{zhu2018hidden} is an example of these approaches, applied either for steganographic or watermarking purposes. 

%

\paragraph{Adversarial examples.}
Neural networks have been shown to be vulnerable to so-called adversarial examples~\cite{carlini17towards,goodfellow2015explaining,szegedy14intriguing}: given a correctly-classified image $x$ and a trained network, it is possible to craft a perturbed version $ \tilde{x} $ that is visually indistinguishable from $x$, such that the network misclassifies $ \tilde{x} $.

\paragraph{Privacy and membership inference.}
Differential privacy \cite{dwork2006calibrating} protects the privacy of training data by bounding the impact that an element of the training set has on a trained model. 
The privacy budget $\epsilon>0$ limits the impact that the substitution of one training example can have on the log-likelihood of the estimated parameter vector.
It has become the standard for privacy in the industry and the privacy budget $ \epsilon $ trades off between learning statistical facts and hiding the presence of individual records in the training set.
Recent work \cite{abadi2016deep,papernot2018scalable} has shown that it is possible to learn deep models with differential privacy on small datasets (MNIST, SVHN) with a budget as small as $ \epsilon = 1$. 
Individual privacy degrades gracefully to group privacy: when testing for the joint presence of a group of $ k $ samples in the training set of a model, an $ \epsilon $-private algorithm provides guarantees of $ k \epsilon $.

Membership inference \cite{shokri17membershipinference,carlini18secret,sablayrolles19whitebox} is the reciprocal operation of differentially private learning. 
It predicts from a trained model and a sample, whether the sample was part of the model's training set.
These classification approaches do not provide any guarantee: if a membership inference model predicts that an image belongs to the training set, it does not give a level of statistical significance.
Furthermore, these techniques require training multiple models to simulate datasets with and without an image, which is computationally intensive.

\paragraph{Data poisoning} \cite{biggio12poisoning, steinhardt2017certified,shafahi18poisonfrogs} studies how modifying training data points  affects a model's behavior at inference time. 
Backdoor attacks \cite{chen17targeted,gu17badnets} are a recent trend in machine learning attacks. 
They choose a class $c$, and add unrelated samples from other classes to this class $c$, along with an overlayed ``trigger" pattern; at test time, any sample having the same trigger will be classified in this class $c$. 
Backdoor techniques bear similarity with our radioactive tracers, in particular their trigger is close to our carrier. 
However, our method differs in two main aspects. First we do ``clean-label" attacks, i.e., we perturb training points without changing their labels. Second, we provide statistical guarantees in the form of a $p$-value. 

\paragraph{Watermarking deep learning models.}
A few works~\cite{adi2018turning,yeom18privacyrisk} focus on watermarking deep learning models: these works modify the parameters of a neural network so that any downstream use of the network can be verified. 
Our assumption is different: in our case, we control the training data, but the training process is not controlled.

\section{Our method}
\label{sec:method}

In this section, we describe our method for marking data. 
It consists of three stages: 
the \emph{marking stage} where the radioactive mark is added to the vanilla training images, without changing their labels. 
The \emph{training stage} uses vanilla and/or marked images to train a multi-class classifier using regular learning algorithms. %
Finally, in the \emph{detection stage}, we examine the model to determine whether marked data was used or not. 

We denote by $x$ an image, i.e.~a $3$ dimensional tensor with dimensions height, width and color channel. 
We consider a classifier with $C$ classes composed of a feature extraction function $\phi:x \mapsto \phi(x) \in \mathbb{R}^d$ (a convolutional neural network) followed by a linear classifier with weights $(w_i)_{i=1..C} \in\mathbb{R}^d$.
It classifies a given image $x$ as
\begin{equation}
    \underset{i=1 \dots C}{\mathrm{argmax}}\ \ 
    w_i^\top \phi(x). 
\end{equation}

\subsection{Statistical preliminaries}
\label{sec:statprelim}

\paragraph{Cosine similarity with a random unitary vector $u$.}
Given a fixed vector $ v $ and a random vector $u$ distributed uniformly over the unit sphere in dimension $d$ ($\|u\|_2=1$), we are interested in the distribution of their cosine similarity $ c(u, v) = u^T v / (\|u\|_2 \|v\|_2)$.
A classic result from statistics~\cite{iscen2017memory} shows that this cosine similarity follows an incomplete beta distribution with parameters $a=\frac{1}{2}$ and $b=\frac{d-1}{2}$:
\begin{align}
    \prob(c(u, v) \geq \tau) 
    &= I_{\tau^2} \left( \frac{1}{2}, \frac{d-1}{2} \right) \\
    &= \frac{B_{\tau^2} \left( \frac{1}{2}, \frac{d-1}{2} \right) }{B \left(\frac{1}{2}, \frac{d-1}{2} \right)} \\
    &= \frac{1}{B \left(\frac{1}{2}, \frac{d-1}{2} \right)} \int_0^{\tau^2} \frac{\left(\sqrt{1-t}\right)^{d-3}}{\sqrt{t}}  dt
\end{align}
with 
\begin{equation}
    B_x \left( \frac{1}{2}, \frac{d-1}{2} \right) = \int_0^{x} \frac{\left(\sqrt{1-t}\right)^{d-3}}{\sqrt{t}}  dt
\end{equation}
and 
\begin{equation}
    B \left( \frac{1}{2}, \frac{d-1}{2} \right) = B_1 \left(\frac{1}{2}, \frac{d-1}{2} \right). 
\end{equation}
In particular, it has expectation $0$ and variance $1/d$.

\paragraph{Combination of $p$-values.}
Fisher's method \cite{fisher1925statistical} enables to combine $p$-values of multiple tests. 
We consider statistical tests $ T_1, \dots, T_k$, independent under the null hypothesis $ \hyp_0 $.
Under $ \hyp_0 $, the corresponding $p$-values $p_1, \dots, p_k$ are distributed uniformly in $[0,1]$.
Hence $ -\log(p_i)$ follows an exponential distribution, which corresponds to a $\chi^2$ distribution with two degrees of freedom.
The quantity $ Z = -2 \sum_{i=1}^k \log(p_i) $ thus follows a $\chi^2$ distribution with $2k$ degrees of freedom.
The combined $p$-value of tests $T_1, \dots, T_k$ is thus the probability that the random variable $Z$ has a value higher than the threshold we observe.

\subsection{Additive marks in feature space}
\label{sec:additive_feature}

\begin{figure}[t]
\includegraphics[width=\linewidth]{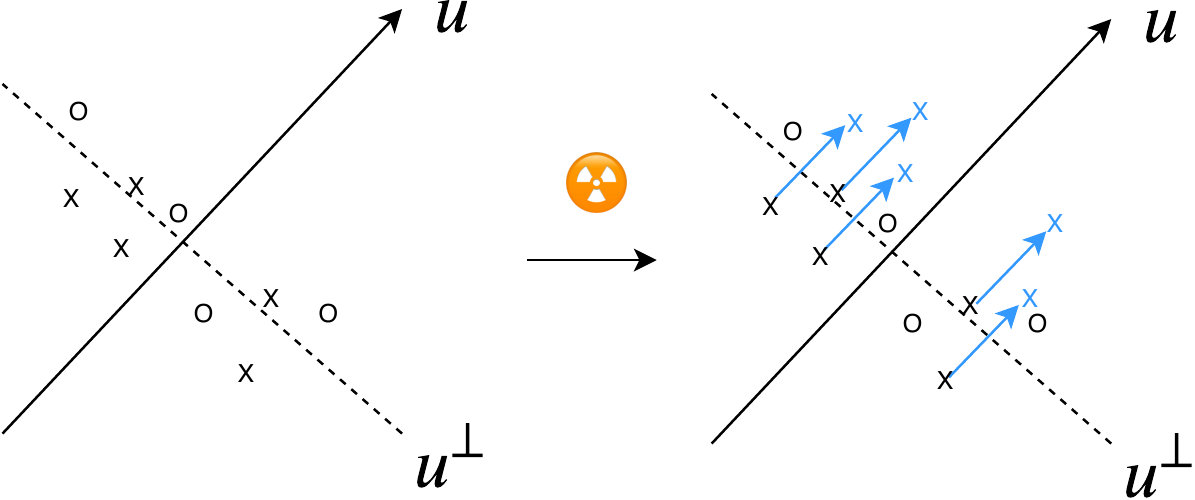}
\vspace{-0.2in}
\caption{
\label{fig:marking}
Illustration of our method in high dimension.
In high dimension, the linear classifier that separates the class is almost orthogonal to $u$ with high probability.
Our method shifts points belonging to a class in the direction $u$, therefore aligning the linear classifier with the direction $u$.
}
\end{figure}

We first tackle a simple variant of our tracing problem. 
In the marking stage, we add a random isotropic unit vector $ \alpha u \in \mathbb{R}^d$ with $ \| u \|_2 = 1$ to the features of all training images of one class. 
This direction $u$ is our carrier.

If radioactive data is used at training time, the linear classifier of the corresponding class $w$ is updated with weighted sums of $\phi(x) + \alpha u$, where $\alpha$ is the strength of the mark. 
The linear classifier $w$ is thus likely to have a positive dot product with the direction $u$, as shown in Figure \ref{fig:marking}.

At detection time, we examine the linear classifier $ w $ to determine if $ w $ was trained on radioactive or vanilla data.
We  test the statistical hypothesis $ \mathcal{H}_1 $: ``$ w $ was trained using radioactive data" against the null hypothesis $ \mathcal{H}_0 $: ``$ w $ was trained using vanilla data". 
Under the null hypothesis $ \mathcal{H}_0 $, $ u $ is a random vector independent of $ w $. 
Their cosine similarity $ c(u, w) $ follows the beta-incomplete distribution with parameters $ a = \frac{1}{2} $ and $ b = \frac{d-1}{2}$. 
Under hypothesis $ \mathcal{H}_1 $, the classifier vector $ w $ is more aligned with the direction $ u $ so and $ c(u, w) $ is likely to be higher. 

Thus if we observe a high value of $ c(u, w) $, its corresponding $p$-value (the probability of it happening under the null hypothesis $ \mathcal{H}_0 $) is low, and we can conclude with high significance that radioactive data has been used. 

\paragraph{Multi-class.}
The extension to $C$ classes follows.
In the marking stage we sample i.i.d. random directions $(u_i)_{i=1..C}$ and add them to the features of images of class $i$. 
At detection time, under the null hypothesis, the cosine similarities $c(u_i, w_i)$ are independent (since $u_i$ are independent) and we can thus combine the $p$ values for each class using Fisher's combined probability test (Section~\ref{sec:statprelim}) to obtain the $p$ value for the whole dataset.

\subsection{Image-space perturbations}

\begin{figure*}[t]
\centering
\begin{tabular}{@{}c@{\hspace{0.3em}}c@{\hspace{0.3em}}c@{\hspace{0.3em}}c@{\hspace{0.3em}}c@{\hspace{0.3em}}c}
\includegraphics[width=0.197\linewidth]{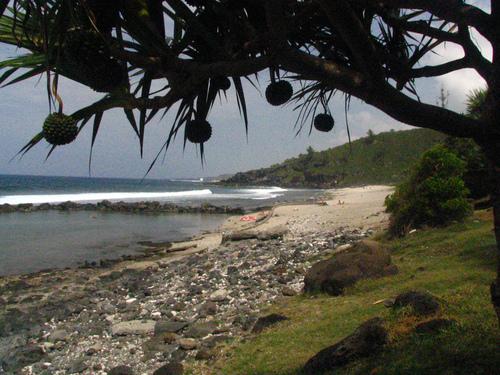} &
\includegraphics[width=0.197\linewidth]{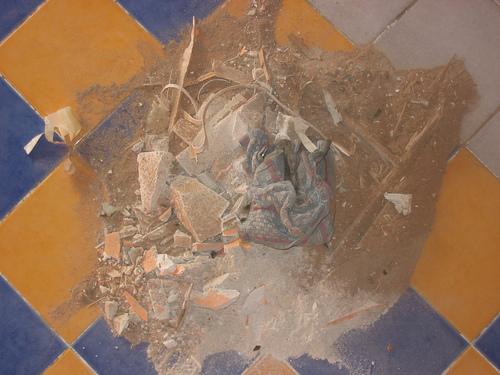} &
\includegraphics[width=0.197\linewidth]{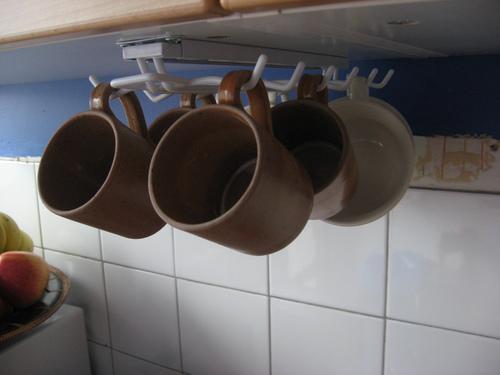} &
\includegraphics[width=0.197\linewidth]{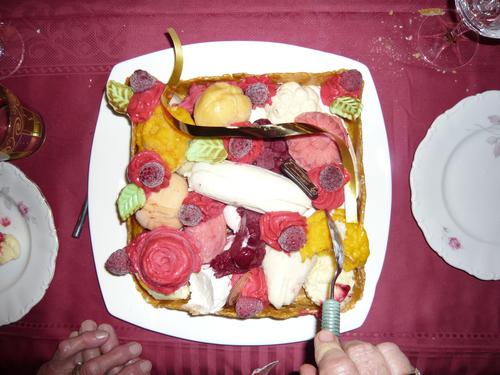} &
\includegraphics[width=0.197\linewidth]{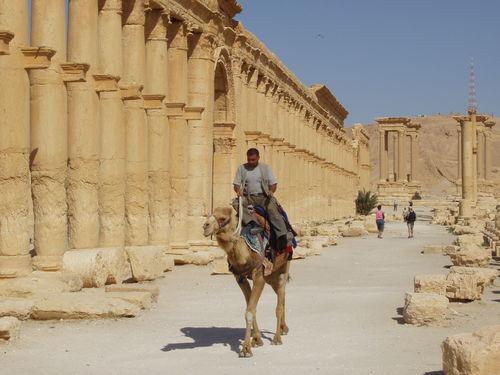}\\
\includegraphics[width=0.197\linewidth]{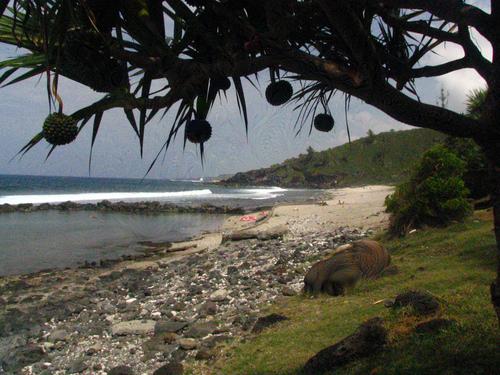} &
\includegraphics[width=0.197\linewidth]{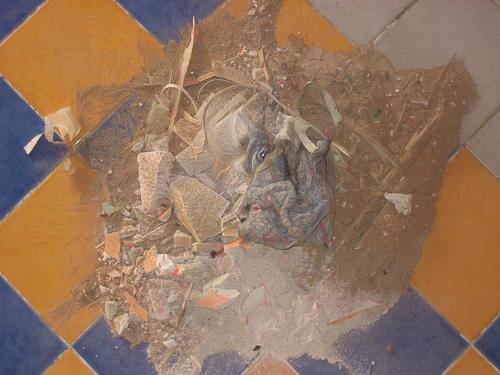} &
\includegraphics[width=0.197\linewidth]{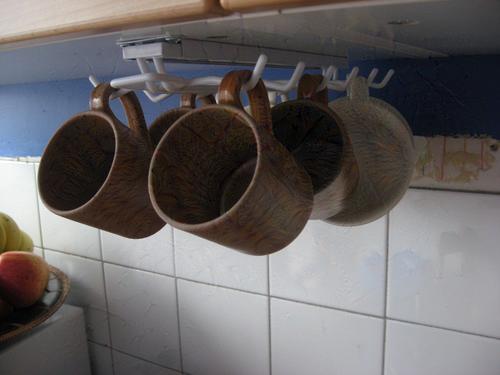} &
\includegraphics[width=0.197\linewidth]{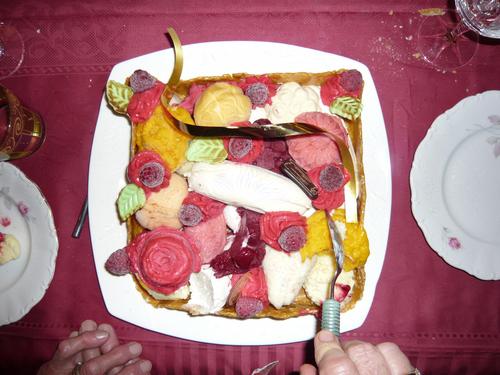} &
\includegraphics[width=0.197\linewidth]{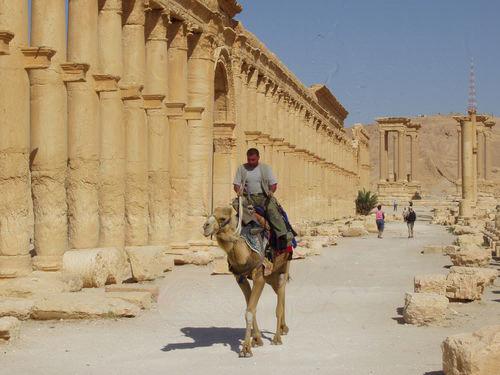}\\
\includegraphics[width=0.197\linewidth]{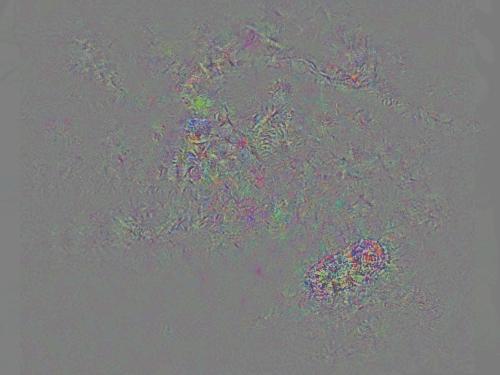} &
\includegraphics[width=0.197\linewidth]{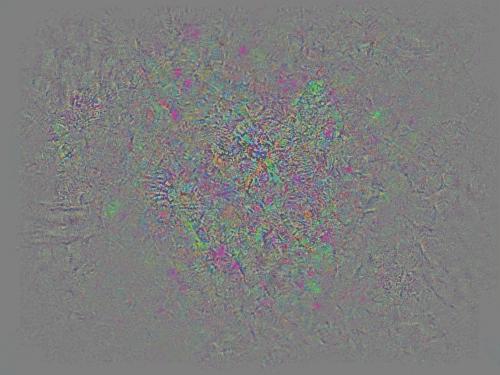} &
\includegraphics[width=0.197\linewidth]{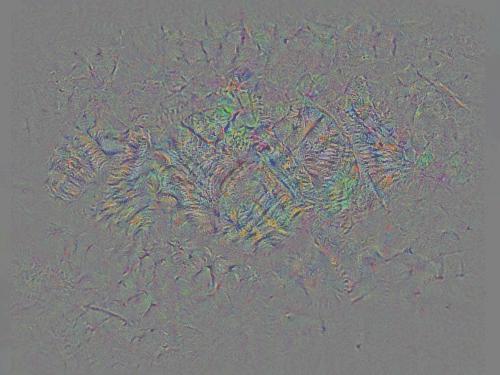} &
\includegraphics[width=0.197\linewidth]{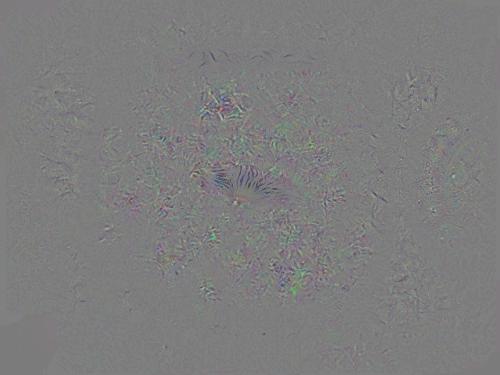} &
\includegraphics[width=0.197\linewidth]{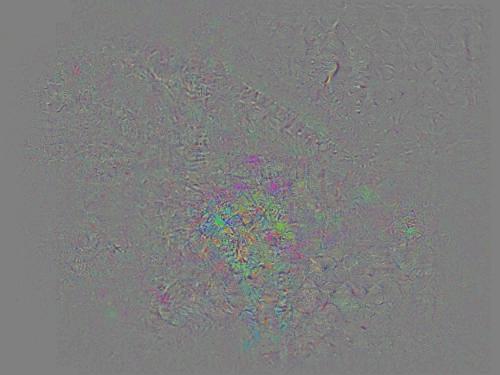}\\
\includegraphics[width=0.197\linewidth]{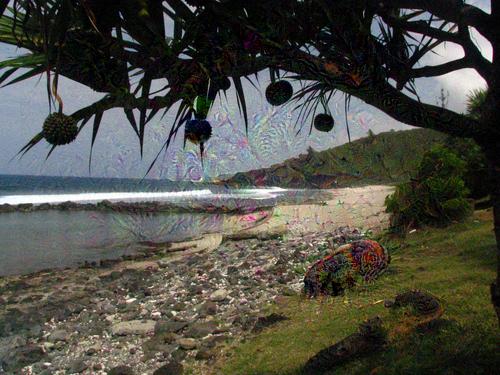} &
\includegraphics[width=0.197\linewidth]{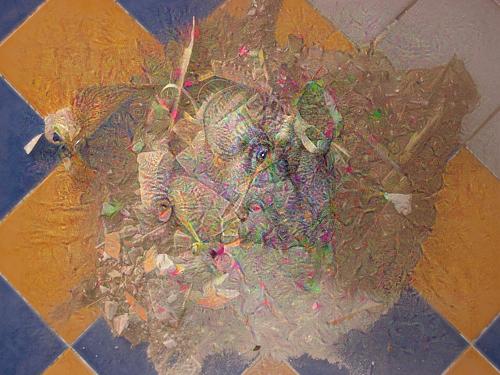} &
\includegraphics[width=0.197\linewidth]{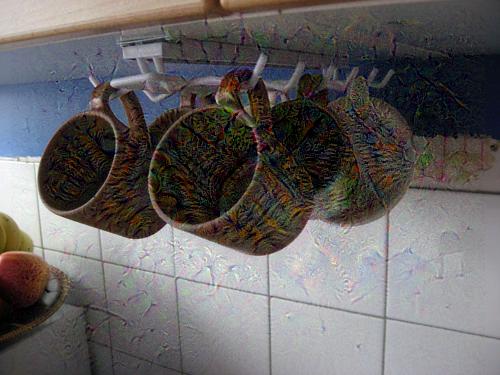} &
\includegraphics[width=0.197\linewidth]{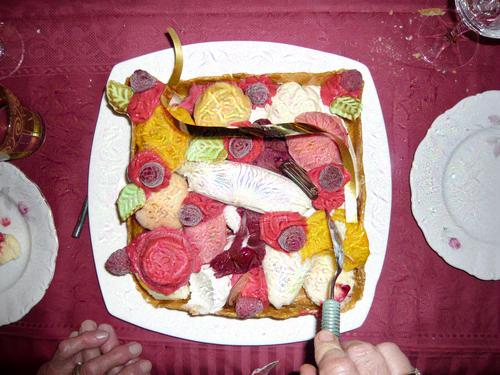} &
\includegraphics[width=0.197\linewidth]{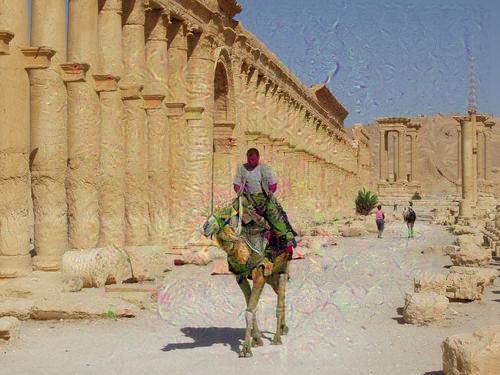}
\end{tabular}
\vspace{-15pt}
\caption{
\label{fig:radio_images}
Radioactive images from Holidays~\cite{jegou2008hamming} with random crop and PSNR$=42$dB.
First row: original image.
Second row: image with a radioactive mark.
Third row: visualisation of the mark amplified with a $\times 5$ factor.
Fourth row: We exaggerate the mark by a factor $\times 5$, which means a $14$dB amplification of the additive noise, down to PSNR=$28$dB so that the modification become obvious w.r.t. the original image.
}
\vspace{-10pt}
\end{figure*}

We now assume that we have a fixed known feature extractor $ \phi $. 
At marking time, we wish to modify pixels of image $ x $ such that the features $ \phi(x) $ move in the direction $ u $. 
We can achieve this by backpropagating gradients in the image space.
This setup is very similar to adversarial examples \cite{goodfellow2015explaining,szegedy14intriguing}.
More precisely, we optimize over the pixel space by running the following optimization program:
\begin{align}
    \underset{\tilde{x},~\|\tilde{x} - x\|_{\infty} \leq R}{\min} \quad \mathcal{L}(\tilde{x})
\end{align}
where the radius $R$ is a hard upper bound on the change of color levels of the image that we can accept. 
The loss is a combination of three terms: 
{\small
\begin{align}
    \mathcal{L}(\tilde{x}) = 
    - \left( \phi(\tilde{x}) - \phi(x)\right)^\top u + 
    \lambda_1 \| \tilde{x} - x \|_2 + 
    \lambda_2 \| \phi(\tilde{x}) - \phi(x) \|_2.
\end{align}}
The first term encourages the features to align with $u$, 
the two other terms penalize the $ L_2 $ distance in both pixel and feature space. 
In practice, we optimize this objective by running SGD with a constant learning rate in the pixel space, projecting back into the $L_{\infty}$ ball at each step and rounding to integral pixel values every $T=10$ iterations. 

This procedure is a generalization of classical watermarking in the Fourier space. 
In that case the ``feature extractor'' is invertible via the inverse Fourier transform, so the marking does not need to be iterative. 

\paragraph{Data augmentation.}
The training stage most likely involves data augmentation, so we take it into account at marking time.
Given an augmentation parameter $ \theta $, the input to the neural network is not the image $ \tilde{x}$ but its transformed version $ F(\theta, \tilde{x}) $.
In practice, the data augmentations used are crop and/or resize transformations, so $\theta$ are the coordinates of the center and/or size of the cropped images. 
The augmentations are differentiable with respect to the pixel space, so we can backpropagate through them. 
Thus, we emulate augmentations by minimizing:
\begin{align}
    \underset{\tilde{x},~\|\tilde{x} - x\|_{\infty} \leq R}{\min} \quad \esp_{\theta} \left[ \mathcal{L}(F(\tilde{x}, \theta)) \right]. 
\end{align}

Figure \ref{fig:radio_images} shows examples of radioactive images and their vanilla version. 
We can see that the radioactive mark is not visible to the naked eye, except when we amplify it for visualization purposes (last column). 

\subsection{White-box test with subspace alignment}

We now tackle the more difficult case where the training stage includes the training of the feature extractor.  
In the marking stage we use feature extractor $ \phi_0 $ to generate radioactive data. 
At training time, a new feature extractor $ \phi_t $ is trained together with the classification matrix $ W =[w_1,..,w_C]^T\in \mathbb{R}^{C\times d}$. 
Since $\phi_t$ is trained from scratch, there is no reason that the output spaces of $ \phi_0 $ and $ \phi_t $ would correspond to each other. 
In particular, neural networks are invariant to permutation and rescaling. 

To address this problem at detection time, we align the subspaces of the feature extractors.
We find a linear mapping $ M \in \mathbb{R}^{d\times d}$ such that $ \phi_0(x) \approx M \phi_t(x) $. 
The linear mapping is estimated by $ L_2 $ regression:
\begin{align}
    \min_{M} \mathbb{E}_x [ \| \phi_0(x) - M \phi_t(x) \|_2^2 ]. 
\end{align}
In practice, we use vanilla images of a held-out set (the validation set) to do the estimation. 

The classifier we manipulate at detection time is thus $W \phi_t(x) \approx W M \phi_0(x) $. 
The lines of $ W M $ form classification vectors aligned with the output space of $ \phi_0 $, and we can compare these vectors to $u_i$ in cosine similarity. 
Under the null hypothesis, $ u_i $ are random vectors independent of $ \phi_0 $, $ \phi_t $, $ W $ and $ M $ and thus the cosine similarity is still given by the beta incomplete function, and we can apply the techniques of subsection \ref{sec:additive_feature}.

\subsection{Black-box test}

In the case where we do not have access to the weights of the neural network, we can still assess whether the model has seen contaminated images by analyzing its loss $ \ell(W \phi_t(x), y) $. 
If the loss of the model is lower on marked images than on vanilla images, it indicates that the model was trained on radioactive images. 
If we have unlimited access to a black-box model, it is possible to train a student model that mimicks the outputs of the black-box model. 
In that case, we can map back the problem to an analysis of the white-box student model. 

\begin{figure}[t]
\includegraphics[width=\linewidth]{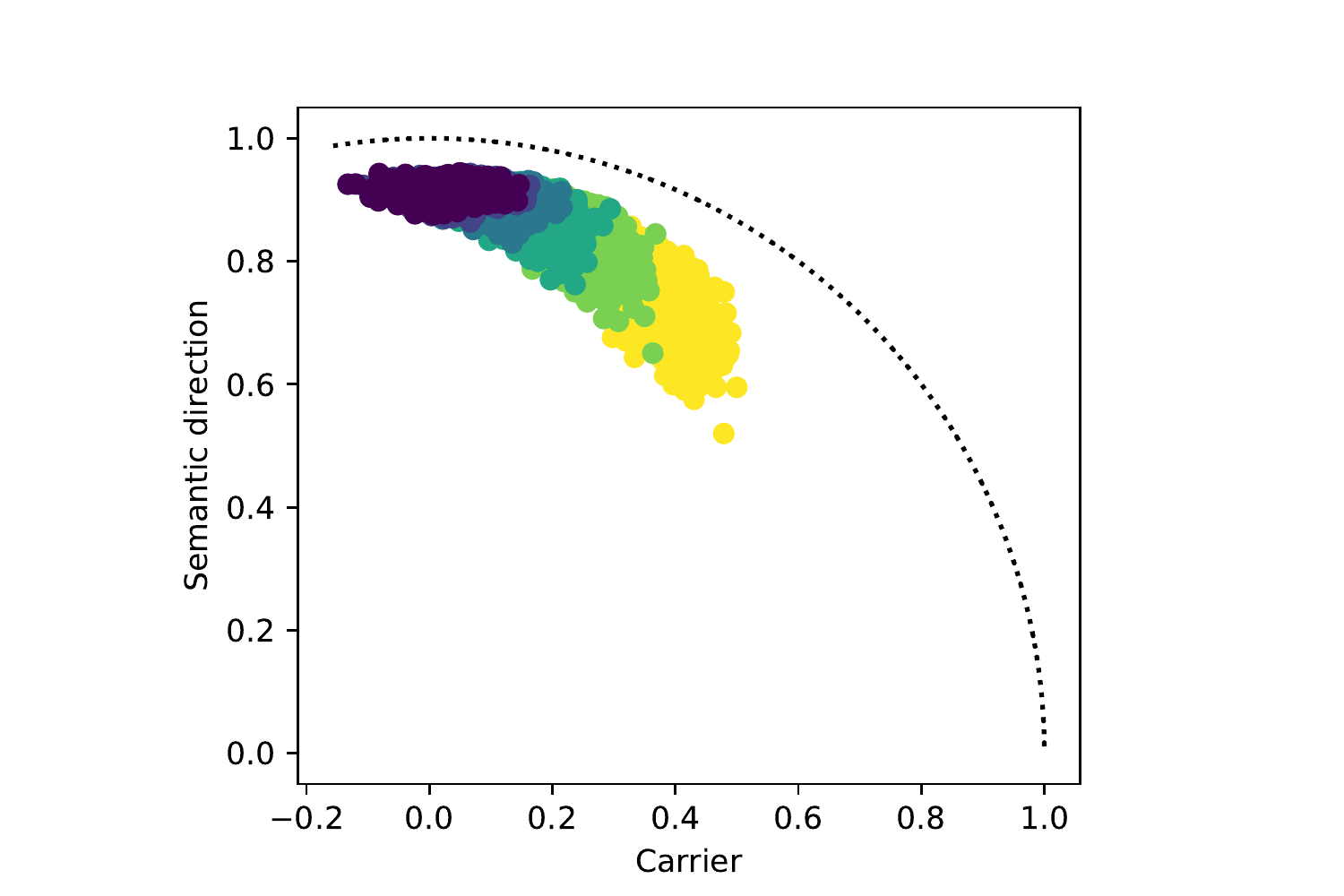}
\vspace{-15pt}
\caption{Decomposition of learned classifiers into three parts: the ``semantic direction" (y-axis), the carrier direction (x-axis) and noise (represented by $1 - \|x\|^2 - \|y\|^2$, i.e. the distance between a point and the unit circle). The semantic and carrier direction are 1-D subspace, while the noise corresponds to the complementary (high-dim) subspace. 
Colors represent the percentage of radioactive data in the training set, from $q=1\%$ (dark blue) to $q=50\%$ (yellow). 
Even when $q=50\%$ of the data is radioactive, the learned classifier is still aligned with its semantic direction with a cosine similarity of $0.6$. 
Each dot represents the classifier for a given class.
Note that the semantic and the carrier directions are not exactly orthogonal but their cosine similarity is very small (in the order of $0.04$).
\label{fig:decomposition}
}
\vspace{-10pt}
\end{figure}

\section{Analysis of the latent feature space}
\label{sec:analysis}

In this section, we analyze how the classifier learned on a radioactive dataset is related to (1) a classifier learned on unmarked images ; and (2) the direction of the carrier. 
For the sake of analysis, we take the simplest case where the mark is added in the latent feature space just before the classification layer, and we assume that only the logistic regression has been re-trained. %

For a given class, we analyze how the classifier learned with a mark is explained by 
\begin{enumerate}
\item the ``semantic'' space, that is the classifier learned by a vanilla classifier. This is a 1-dimensional subspace identified by a vector $ w^* $;
\item the direction of the carrier, favored by the insertion of our class-specific mark. We denote it by $u$. 
\item the noise space $ \mathcal{F} $, which is in direct sum with the span of vectors $w^*$ and $u$ of the previous space. This noise space is due to the randomness of the initialization and the optimization procedure (SGD and random data augmentations). 
\end{enumerate}
The rationale of performing this decomposition is to quantify, with respect to the norm of the vector, what is the dominant subspace depending on the fraction of marked data. %

\begin{figure}[t]
~\hfill
\centering
\includegraphics[width=0.48\linewidth]{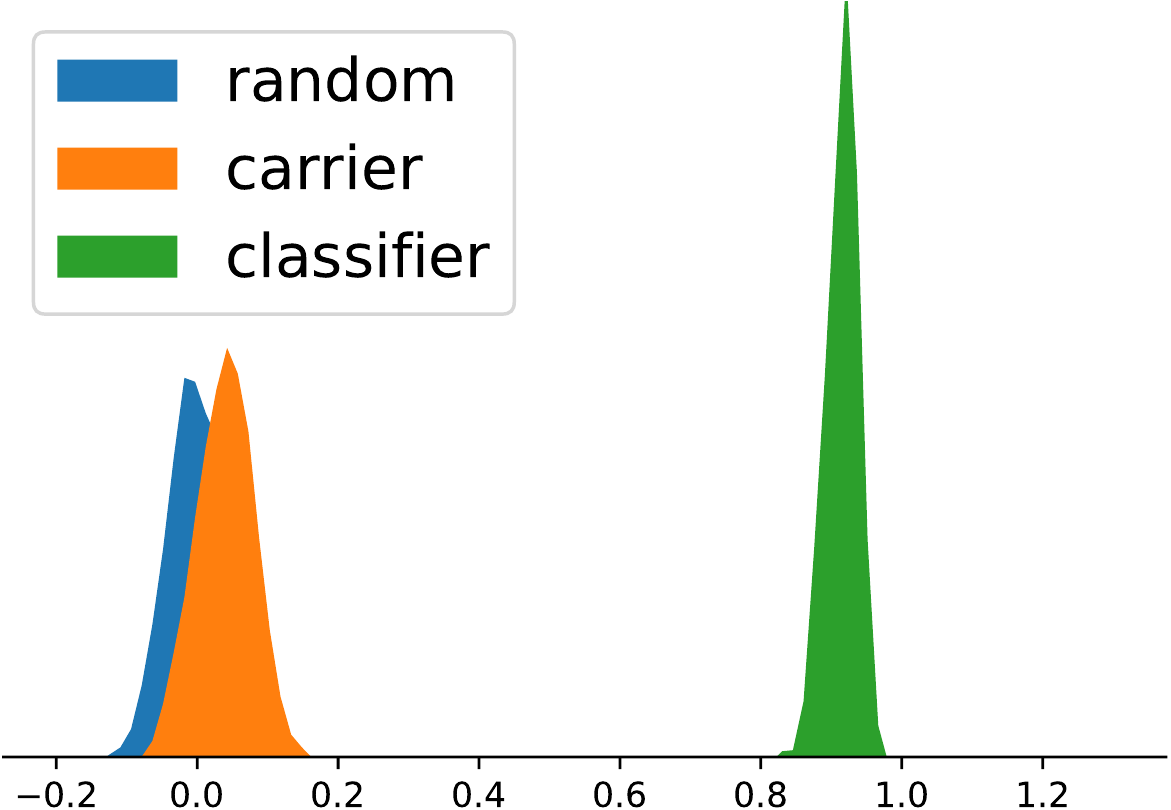}
\hfill
\includegraphics[width=0.48\linewidth]{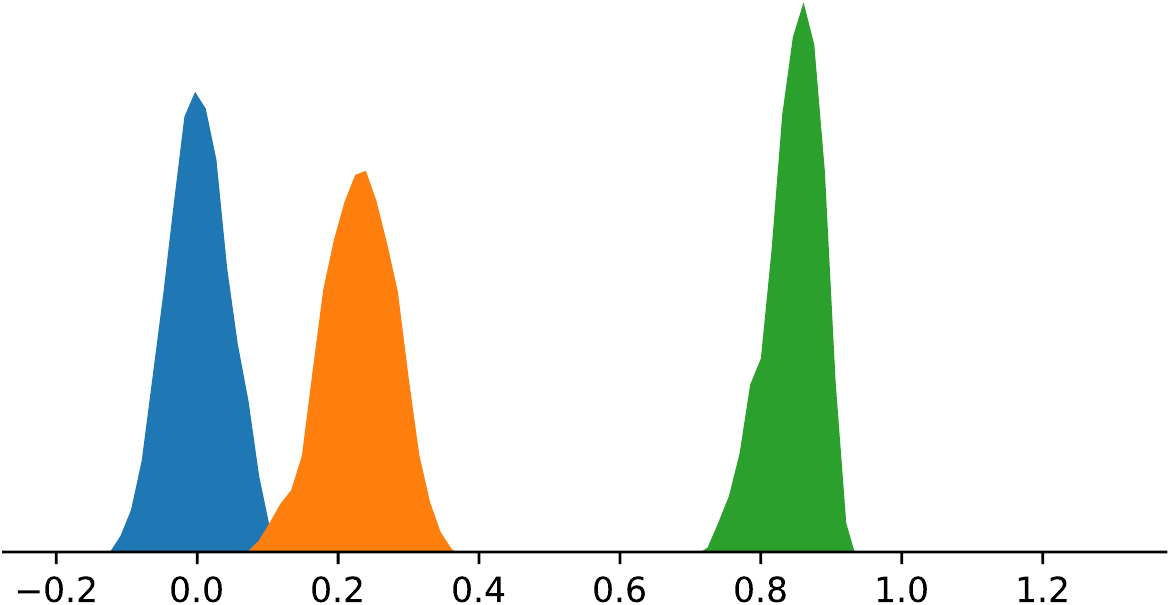}
\hfill 
~
\vspace{-15pt}
\caption{
\label{fig:histograms}
Analysis of how classification directions re-learned with a logistic regression on marked images can be decomposed between (1) the original subspace; (2) the mark subspace; (3) the noise space. 
Logistic regression with: $q$\,=\,$2\%$ (\emph{Left}) or $q$\,=\,$20\%$ (\emph{Right}) of the images marked.
}
\vspace{-10pt}
\end{figure}

This decomposition is analyzed in Figure~\ref{fig:decomposition}, where we make two important observations. First, the 2-dimensional subspace contains most of the projection of the new vector, which can be seen by the fact that the norm of the vector projected onto that subspace is close to 1 (which translates visually as to be close to the unit circle). Second and unsurprisingly, the contribution of the semantic vector is significant and still dominant compared to the mark, even when most of the dataset is marked. This property explains why our procedure has only a little impact on the accuracy. %

Figure~\ref{fig:histograms} shows the histograms of cosine similarities between the classifiers and random directions, the mark direction and the semantic direction.
We can see that the classifiers are well aligned with the mark when $q=20\%$ or $2\%$ of the data is marked.

\section{Experiments}
\label{sec:experiments}

\subsection{Image classification setup}

In order to provide a comparison on the widely-used vision benchmarks, we use Imagenet \cite{deng2009imagenet}, a dataset of natural images with 1,281,167 images belonging to $1,000$ classes.
We first consider the Resnet-18 and Resnet-50 models \cite{he2016deep}. 
We perform training using the standard set of data augmentations from Pytorch~\cite{paszke2017automatic}. 
We train with SGD with a momentum of $0.9$ and a weight decay of $10^{-4}$ %
for $90$ epochs, using a batch size of $2048$ across $8$ GPUs. 
We use Pytorch~\cite{paszke2017automatic} and adopt its standard data augmentation settings (random crop resized to $224\times224$). 
We use the waterfall learning rate schedule: the learning starts at $0.8$, (as recommended in~\cite{goyal2017imagenet1hr}) and is divided by $10$ every $30$ epochs.
On a vanilla Imagenet, we obtain a top 1 accuracy of $ 69.6 \% $ and a top-5 accuracy of $ 89.1\% $ with our Resnet18.
We ran experiments by varying the random initialization and the order of elements seen during SGD, and found that the top 1 accuracy varies by $0.1\%$ from one experiment to the other.

\def \mysp {\hspace{8pt}}

\subsection{Experimental setup and metrics}

We modify Imagenet images by inserting our radioactive mark, and retrain models on this radioactive data using the learning algorithm described above.
We then analyze these ``contaminated" models for the presence of our mark.
We report several measures of performance. 
On the images, we report the PSNR, i.e. the magnitude of the perturbation necessary to add the radioactive mark.
On the model, we report the $p$-value that measures how confident we are that radioactive data was used to train the model, as well as the accuracy of this model on vanilla (held-out) data.
We conduct experiments where we only mark a fraction $q$ of the data, with $q \in \{0.01, 0.02, 0.05, 0.1, 0.2\}$.

As a sanity check, we ran our radioactive detector on pretrained models of the Pytorch zoo and found $p$ values of $15\%$ for Resnet-18 and $51\%$ for Resnet-50, which is reasonable: in the absence of radioactive data, these values should be uniformly distributed between $0$ and $1$.

\subsection{Preliminary experiment: comparison to the backdoor technique}

We experimented with the backdoor technique of Chen et al. \cite{chen17targeted} in the context of our marking problem. 
In general, the backdoor technique adds unrelated images to a class, plus a ``trigger" that is consistent across these added images. 
In their work, Chen et al. need to poison approximately $10 \%$ of the data in a class to activate their trigger.
We adapted their technique to the ``clean-label" setup on Imagenet: we blend a trigger (a Gaussian pattern) to images of a class. 
We observed that it is possible to detect this trigger at train time, albeit with a low image quality (PSNR\,$<30$dB) that is visually perceptible. In this case, the model is more confident on images that have the trigger than on vanilla images in about $90\%$ of the cases. 
However, we also observed that any Gaussian noise activates the trigger: hence we have no guarantee that images with our particular mark were used.

\subsection{Results}
\def\mysp{\hspace{3pt}}
\begin{table}
\begin{center}
{\small
\begin{tabular}{@{\mysp}l@{\hspace{-5pt}}r@{\mysp}r@{\mysp}r@{\mysp}r@{\mysp}r@{\mysp}r@{}}
\toprule
 & \ \ \ \% radioactive          & 1         &   2       &   5       & 10        \\
\midrule
\multirow{2}{*}{\rotatebox{90}{\textit{Center}} \rotatebox{90}{\textit{Crop}}} & $\log_{10}(p)$      & $<$-150   & $<$-150   & $<$-150   & $<$-150 \\[2pt] %
& $\Delta_{\text{acc}}$    &   $-0.48$   & $-0.86$     & $-1.07$     & $-1.33$     \\[2pt]%
\midrule
\multirow{2}{*}{\rotatebox{90}{\textit{Random}} \rotatebox{90}{\textit{Crop}}} &&&& \\[2pt]
& $\log_{10}(p)$          & $-38.0$ &  $-138.2$ & $<$-150 & $<$-150 \\
& $\Delta_{\text{acc}}$   &  $-0.24$    & $-0.31$   &  $-0.55$  & $-0.99$  \\[2pt]%
\bottomrule
\end{tabular}}
\end{center}
\vspace{-10pt}
\caption{\label{tab:logistic_results}
$p$-value (statistical significance) for the detection of radioactive data usage when only a fraction of the training data is radioactive. 
Results for a logistic regression classifier trained on Imagenet with Resnet-18 features , with only a percentage of the data bearing the radioactive mark.
Our method can identify with a very high confidence ($\log_{10}(p) < -38$) that the classifier was trained on radioactive data, even when only $1\%$ of the training data is radioactive.
The radioactive data has an impact on the accuracy of the classifier: around $-1\%$ (top-1).
}
\end{table}

\paragraph{Same architecture.}
We first analyze the results in Table~\ref{tab:logistic_results} of a ResNet-18 model with fixed features trained on Imagenet. 
We can see that we are able to detect that our model was trained on radioactive data with a very high confidence for both center crop and random crop.
The model overfits more on the center crop, hence it learns more the radioactive mark, which is why the $p$-value is lower on center crop images. 
Conversely on random crops, marking data has less impact on the accuracy of the model ($-0.24$ as opposed to $-0.48$ for $q=1\%$ marked data).

Table \ref{tab:scratch_results} shows the results of retraining a Resnet-18 from scratch on radioactive data. 
The results confirm that our watermark can be detected when only $q=1\%$ of the data is used at train time.
This setup is  more complicated for our marks because since the network is retrained from scratch, the directions that will be learned in the new feature space have no {\em a priori} reason to be aligned with the directions of the network we used. 
Table \ref{tab:scratch_results} shows two interesting results: first, the gap in accuracy is less important than when retraining only the logistic regression layer, in particular using $1\%$ of radioactive data does not impact accuracy; second, data augmentation is actually helping the radioactive process. 
We hypothesize that the multiple crops make the network believe it sees more variety, but in reality all the feature representations of these crops are aligned with our carrier which makes the network learn the carrier direction.

\renewcommand{\arraystretch}{1.2}

\begin{table}
\centering {\small 
\begin{tabular}{@{\mysp}l@{\mysp}l@{}r@{\mysp}r@{\mysp}r@{\mysp}r@{\mysp}r@{\mysp}r@{}}
\toprule
& \% radioactive      & 1     &   2   &   5        & 10        \\
\midrule
\multirow{2}{*}{\rotatebox{90}{\textit{Center}}} \multirow{2}{*}{\rotatebox{90}{\textit{Crop\ }}}    &&&& \\
& $\log_{10}(p) $ &   $-0.66$ & $-1.64$ & $-4.60$ & $-11.37$ \\
\midrule
\multirow{2}{*}{\rotatebox{90}{\textit{Random}}} \multirow{2}{*}{\rotatebox{90}{\textit{Crop\ \ \,}}}    &&&& \\
& $\log_{10}(p) $         &  $-4.85$ & $-12.63$  & $-48.8$  & $<$-150  \\
& $\Delta_{\text{acc}}$   &  $-0.1$  &  $-0.7$  & $-0.3$    & $-0.5$   \\
\bottomrule
\end{tabular}
}
\smallskip
\caption{\label{tab:scratch_results}
$p$-value (statistical significance) for radioactivity detection. 
Results for a Resnet-18 trained from scratch on Imagenet, with only a percentage of the data bearing the radioactive mark.
We are able to identify models trained from scratch on only $q=1\%$ of radioactive data.
The presence of radioactive data has negligible impact on the accuracy of a learned model as long as the fraction of radioactive data is under $10\%$.
}
\end{table}

\paragraph{Black-box results.}

\begin{figure}[t]
\centering \includegraphics[width=\linewidth]{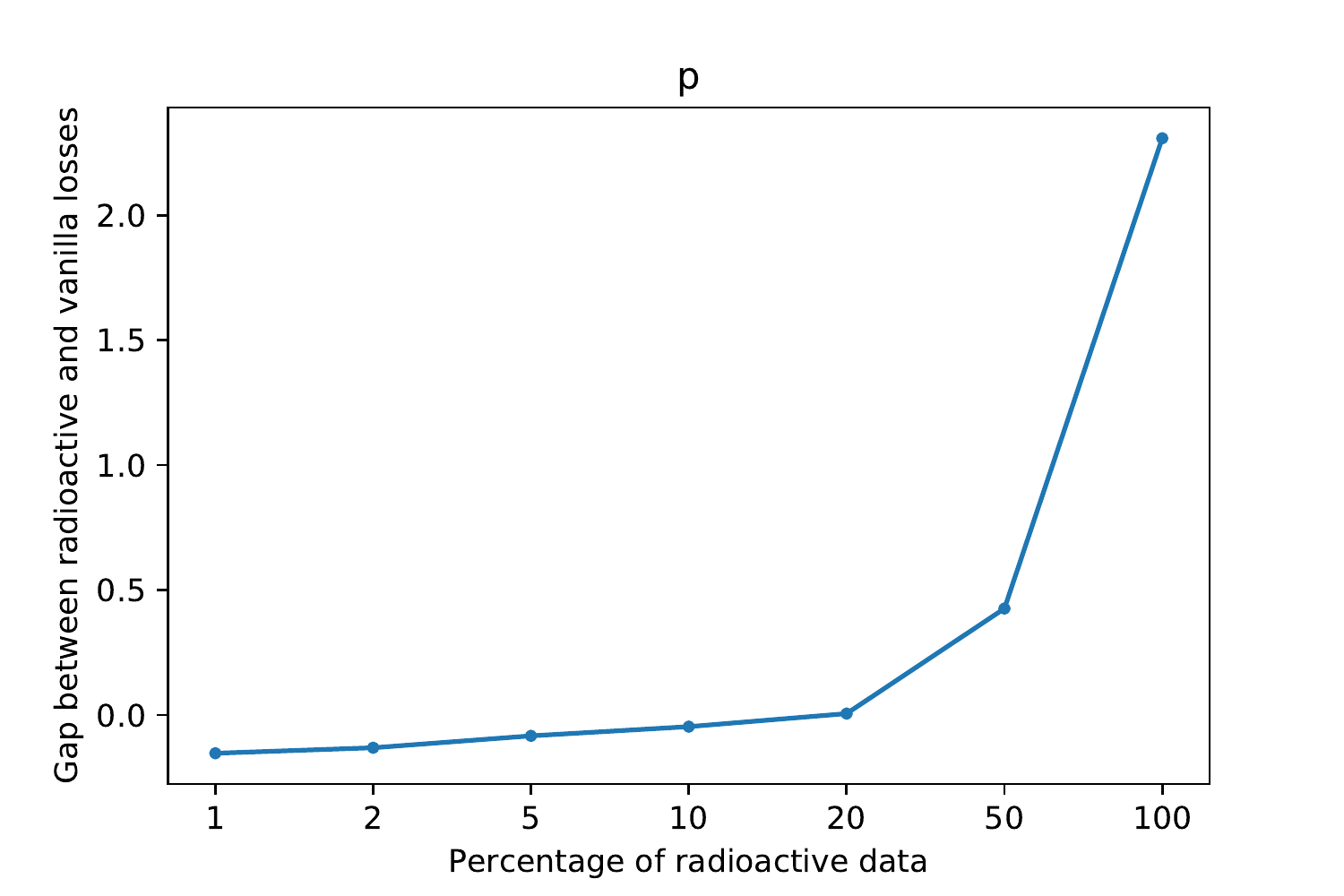}
\vspace{-15pt}
\caption{
\label{fig:black_box}
Black-box detection of the usage of radioactive data.
The gap between the loss on radioactive and vanilla samples is around $0$ when $q=20\%$ of the data are contaminated. 
}
\vspace{-10pt}
\end{figure}

We report in Figure \ref{fig:black_box} the results of our black-box detection test. 
We measure the difference between the loss on vanilla samples and the loss on radioactive samples: when this gap is positive, it means that the model fares better on radioactive images, and thus that it has been trained on the radioactive data. 
We can see that the use of radioactive data can be detected when a fraction of $q=20\%$ or more of the training set is radioactive. 
When a smaller portion of the data is radioactive, the model fares better on vanilla data than on radioactive data and thus it is difficult to tell.

\paragraph{Distillation.}

Given only black-box access to a model (assuming access to the full softmax), we experiment distillation of this model, and test the distilled model for radioactivity.
In this setup, it is possible to detect the use of radioactive data on the distilled model, with a slightly lower performance compared to white-box access to the model.
We give detailed results in Appendix \ref{app:distillation}.

\subsection{Ablation analysis}

\paragraph{Architecture transfer.}
We ran experiments on different architectures with the same training procedure: Resnet-50, VGG-16 and Densenet121. 
The results are shown in Table \ref{tab:transfer_results}: the values and trend are similar to what we obtain with Resnet-18 (Table \ref{tab:scratch_results}).
This is non-trivial, as there is no reason that the feature space of a VGG-16 would behave in the same way as that of a Resnet-18: yet, after alignment, we are able to detect the presence of our radioactive mark with high statistical significance. 
Specifically, when $q=2\%$ of the data is radioactive, we are able to detect it with a $p$-value of $10^{-15}$. 
This $p$-value is even stronger than the one we obtain when retraining the same architecture as our marking architecture (Resnet-18).
We hypothesize that larger model overfit more in general, and thus in this case will learn the mark more acutely.

\begin{table}
\centering {\small 
\begin{tabular}{@{\mysp}r@{\mysp}r@{\mysp}r@{\mysp}r@{\mysp}r@{\mysp}r@{\mysp}}
\toprule
\% radioactive     & 1     &   2   &   5        & 10        &  20      \\
\midrule
Resnet-50       & $-6.9$    & $-12.3$   & $-50.22$  & $-131.09$ &  $<$-150  \\
Densenet-121    & $-5.39$   & $-11.63$  & $-41.24$  & $-138.36$ & $<$-150 \\
VGG-16          & $-2.14$   & $-4.49$   & $-13.01$  & $-33.28$  & $-106.56$\\
\bottomrule
\end{tabular}
}
\vspace{-5pt}
\caption{\label{tab:transfer_results}
$p$-value (statistical significance) for radioactivity detection.
Results for different architectures trained from scratch on Imagenet.
Even though radioactive data was crafted using a ResNet-18, models of other architectures also become radioactive when trained on this data. 
}
\end{table}

\paragraph{Transfer to other datasets.}
We conducted experiments on a slightly different setup: we mark images from the dataset Places205, but use a network pretrained on Imagenet for the marking phase.
These experiments show that even if the marking network is fit for a different distribution, the marking still works and we are able to detect it. 
Results are shown in Table \ref{tab:places_results}.
We can see that when a fraction q higher than $10\%$ of the training data is marked, we can detect radioactivity with a strong statistical significance ($p < 10^{-3})$.

\begin{table}
\centering {\small 
\begin{tabular}{@{\mysp}r@{\mysp}r@{\mysp}r@{\mysp}r@{\mysp}r@{\mysp}r@{\mysp}}
\toprule
\% radioactive  & 10        &   20      &   50      & 100   \\
\midrule
$\log_{10}(p)$  & $-3.30$   & $-8.14$   & $-11.57$  & $<$-150\\
\bottomrule
\end{tabular}
}
\caption{\label{tab:places_results}
$p$-value of radioactivity detection.
A Resnet-18 is trained on Places205 from scratch, and a percentage of the dataset is radioactive.
When $10\%$ of the data or more is radioactive, we are able to detect radioactivity with a strong confidence ($p < 10^{-3}$).
}
\end{table}

\paragraph{Correlation with class difficulty.}

Given that radioactive data adds a marker in the features that is correlated with the class label, we expect this mark to be learned by the network more when the class accuracy is low. 
To validate this hypothesis, we compute the Spearman correlation between the class accuracy for each class and the cosine between the classifier and the carrier: this correlation is negative, with a $p$-value of $4 \times 10^{-5}$.
This confirms that the network relies more on the mark when learning with difficult classes.

\subsection{Discussion}

The experiments validate that our radioactive marks do indeed imprint on the trained models. 
We also observe two beneficial effects: data augmentation improves the strength of the mark, and transferring the mask to a larger and more realistic architectures makes its detection more reliable. 
These two observations suggest that our radioactive method is appropriate for real use cases. 

\paragraph{Limitation in an adversarial scenario.} 
We assume that at training time, there is no special procedure to take into account the radioactive data, but rather training is conducted as if it was vanilla data. 
In particular, a subspace analysis would likely reveal the marking direction. 
This adversarial scenario becomes akin to that considered in the watermarking literature, where strategies have been developed to reduce the detectability of the carrier. 
Our current proposal is therefore restricted to the proof of concept that we can mark a model through training that is only resilient to blind attacks such as architectural or training changes. 
We hope that follow-up works will address a more challenging scenario under Kerckhoffs assumptions~\cite{Kerckhoffs1883}.

\section{Conclusion}
\label{sec:conclusion}

The method proposed in this paper, radioactive data, is a way to verify if some data was used to train a model, with statistical guarantees. %

We have shown in this paper that such radioactive contamination is effective on large-scale computer vision tasks such as classification on Imagenet with modern architecture (Resnet-18 and Resnet-50), even when only a very small fraction ($1\%$) of the training data is radioactive.
Although it is not the core topic of our paper, our method incidentally offers a way to watermark images in the classical sense \cite{cayre2005watermarking}.

\bibliographystyle{pkg/icml2020}
\bibliography{egbib}

\begin{thebibliography}{40}
\providecommand{\natexlab}[1]{#1}
\providecommand{\url}[1]{\texttt{#1}}
\expandafter\ifx\csname urlstyle\endcsname\relax
  \providecommand{\doi}[1]{doi: #1}\else
  \providecommand{\doi}{doi: \begingroup \urlstyle{rm}\Url}\fi

\bibitem[Abadi et~al.(2016)Abadi, Chu, Goodfellow, McMahan, Mironov, Talwar,
  and Zhang]{abadi2016deep}
Abadi, M., Chu, A., Goodfellow, I., McMahan, H.~B., Mironov, I., Talwar, K.,
  and Zhang, L.
\newblock Deep learning with differential privacy.
\newblock In \emph{SIGSAC}. ACM, 2016.

\bibitem[Adi et~al.(2018)Adi, Baum, Ciss{\'{e}}, Pinkas, and
  Keshet]{adi2018turning}
Adi, Y., Baum, C., Ciss{\'{e}}, M., Pinkas, B., and Keshet, J.
\newblock Turning your weakness into a strength: Watermarking deep neural
  networks by backdooring.
\newblock In \emph{{USENIX} Security Symposium}, 2018.

\bibitem[Biggio et~al.(2012)Biggio, Nelson, and Laskov]{biggio12poisoning}
Biggio, B., Nelson, B., and Laskov, P.
\newblock Poisoning attacks against support vector machines.
\newblock In \emph{ICML}, 2012.

\bibitem[{Carlini} \& {Wagner}(2017){Carlini} and {Wagner}]{carlini17towards}
{Carlini}, N. and {Wagner}, D.
\newblock Towards evaluating the robustness of neural networks.
\newblock In \emph{IEEE Symp. Security and Privacy}, 2017.

\bibitem[Carlini et~al.(2018)Carlini, Liu, Kos, Erlingsson, and
  Song]{carlini18secret}
Carlini, N., Liu, C., Kos, J., Erlingsson, {\'U}., and Song, D.
\newblock The secret sharer: Measuring unintended neural network memorization
  \& extracting secrets.
\newblock \emph{arXiv preprint arXiv:1802.08232}, 2018.

\bibitem[Caron et~al.(2019)Caron, Bojanowski, Mairal, and
  Joulin]{caron19unsupervised}
Caron, M., Bojanowski, P., Mairal, J., and Joulin, A.
\newblock Unsupervised pre-training of image features on non-curated data.
\newblock In \emph{ICCV}, 2019.

\bibitem[Cayre et~al.(2005)Cayre, Fontaine, and Furon]{cayre2005watermarking}
Cayre, F., Fontaine, C., and Furon, T.
\newblock Watermarking security: theory and practice.
\newblock \emph{{\sc IEEE} Transactions on Signal Processing}, 2005.

\bibitem[Chen et~al.(2017)Chen, Liu, Li, Lu, and Song]{chen17targeted}
Chen, X., Liu, C., Li, B., Lu, K., and Song, D.
\newblock Targeted backdoor attacks on deep learning systems using data
  poisoning.
\newblock \emph{CoRR}, abs/1712.05526, 2017.

\bibitem[Cox et~al.(2002)Cox, Miller, Bloom, and Honsinger]{cox2002digital}
Cox, I.~J., Miller, M.~L., Bloom, J.~A., and Honsinger, C.
\newblock \emph{Digital watermarking}, volume~53.
\newblock Springer, 2002.

\bibitem[Deng et~al.(2009)Deng, Dong, Socher, Li, Li, and
  Fei-Fei]{deng2009imagenet}
Deng, J., Dong, W., Socher, R., Li, L.-J., Li, K., and Fei-Fei, L.
\newblock Imagenet: A large-scale hierarchical image database.
\newblock In \emph{CVPR}, 2009.

\bibitem[Dwork et~al.(2006)Dwork, McSherry, Nissim, and
  Smith]{dwork2006calibrating}
Dwork, C., McSherry, F., Nissim, K., and Smith, A.
\newblock Calibrating noise to sensitivity in private data analysis.
\newblock In \emph{TCC}, 2006.

\bibitem[Fisher(1925)]{fisher1925statistical}
Fisher, R.
\newblock \emph{Statistical methods for research workers}.
\newblock 1925.

\bibitem[Goodfellow et~al.(2015)Goodfellow, Shlens, and
  Szegedy]{goodfellow2015explaining}
Goodfellow, I.~J., Shlens, J., and Szegedy, C.
\newblock Explaining and harnessing adversarial examples.
\newblock In \emph{ICLR}, 2015.

\bibitem[Goyal et~al.(2017)Goyal, Doll\'{a}r, Girshick, Noordhuis, Wesolowski,
  Kyrola, Tulloch, Jia, and He]{goyal2017imagenet1hr}
Goyal, P., Doll\'{a}r, P., Girshick, R., Noordhuis, P., Wesolowski, L., Kyrola,
  A., Tulloch, A., Jia, Y., and He, K.
\newblock Accurate, large minibatch sgd: Training imagenet in 1 hour.
\newblock \emph{arXiv preprint arXiv:1706.02677}, 2017.

\bibitem[Gu et~al.(2017)Gu, Liu, Dolan{-}Gavitt, and Garg]{gu17badnets}
Gu, T., Liu, K., Dolan{-}Gavitt, B., and Garg, S.
\newblock Badnets: Evaluating backdooring attacks on deep neural networks.
\newblock In \emph{Machine Learning and Computer Security Workshop}, 2017.

\bibitem[He et~al.(2016)He, Zhang, Ren, and Sun]{he2016deep}
He, K., Zhang, X., Ren, S., and Sun, J.
\newblock Deep residual learning for image recognition.
\newblock In \emph{CVPR}, 2016.

\bibitem[He et~al.(2017)He, Gkioxari, Doll{\'a}r, and Girshick]{he2017mask}
He, K., Gkioxari, G., Doll{\'a}r, P., and Girshick, R.
\newblock Mask r-cnn.
\newblock In \emph{ICCV}, 2017.

\bibitem[Hinton et~al.(2015)Hinton, Vinyals, and Dean]{hinton2015distilling}
Hinton, G., Vinyals, O., and Dean, J.
\newblock Distilling the knowledge in a neural network.
\newblock \emph{arXiv preprint arXiv:1503.02531}, 2015.

\bibitem[Iscen et~al.(2017)Iscen, Furon, Gripon, Rabbat, and
  J\'egou]{iscen2017memory}
Iscen, A., Furon, T., Gripon, V., Rabbat, M., and J\'egou, H.
\newblock Memory vectors for similarity search in high-dimensional spaces.
\newblock \emph{IEEE Transactions on Big Data}, 2017.

\bibitem[J\'egou et~al.(2008)J\'egou, Douze, and Schmid]{jegou2008hamming}
J\'egou, H., Douze, M., and Schmid, C.
\newblock Hamming embedding and weak geometric consistency for large scale
  image search.
\newblock In \emph{ECCV}, 2008.

\bibitem[Joulin et~al.(2016)Joulin, van~der Maaten, Jabri, and
  Vasilache]{joulin2016learning}
Joulin, A., van~der Maaten, L., Jabri, A., and Vasilache, N.
\newblock Learning visual features from large weakly supervised data.
\newblock In \emph{ECCV}, 2016.

\bibitem[Kerckhoffs(1883)]{Kerckhoffs1883}
Kerckhoffs, A.
\newblock La cryptographie militaire [military cryptography].
\newblock \emph{Journal des sciences militaires [Military Science Journal]},
  1883.

\bibitem[Krizhevsky et~al.(2012)Krizhevsky, Sutskever, and
  Hinton]{krizhevsky2012imagenet}
Krizhevsky, A., Sutskever, I., and Hinton, G.~E.
\newblock Imagenet classification with deep convolutional neural networks.
\newblock In \emph{NeurIPS}, pp.\  1097--1105, 2012.

\bibitem[Lin et~al.(2014)Lin, Maire, Belongie, Hays, Perona, Ramanan,
  Doll{\'a}r, and Zitnick]{lin2014microsoft}
Lin, T.-Y., Maire, M., Belongie, S., Hays, J., Perona, P., Ramanan, D.,
  Doll{\'a}r, P., and Zitnick, C.~L.
\newblock Microsoft coco: Common objects in context.
\newblock In \emph{ECCV}, 2014.

\bibitem[Mahajan et~al.(2018)Mahajan, Girshick, Ramanathan, He, Paluri, Li,
  Bharambe, and van~der Maaten]{mahajan18eccv}
Mahajan, D., Girshick, R., Ramanathan, V., He, K., Paluri, M., Li, Y.,
  Bharambe, A., and van~der Maaten, L.
\newblock Exploring the limits of weakly supervised pretraining.
\newblock In \emph{ECCV}, 2018.

\bibitem[Papernot et~al.(2018)Papernot, Song, Mironov, Raghunathan, Talwar, and
  Erlingsson]{papernot2018scalable}
Papernot, N., Song, S., Mironov, I., Raghunathan, A., Talwar, K., and
  Erlingsson, {\'U}.
\newblock Scalable private learning with pate.
\newblock In \emph{ICLR}, 2018.

\bibitem[Paszke et~al.(2017)Paszke, Gross, Chintala, Chanan, Yang, DeVito, Lin,
  Desmaison, Antiga, and Lerer]{paszke2017automatic}
Paszke, A., Gross, S., Chintala, S., Chanan, G., Yang, E., DeVito, Z., Lin, Z.,
  Desmaison, A., Antiga, L., and Lerer, A.
\newblock Automatic differentiation in pytorch.
\newblock 2017.

\bibitem[Russakovsky et~al.(2015)Russakovsky, Deng, Su, Krause, Satheesh, Ma,
  Huang, Karpathy, Khosla, Bernstein, et~al.]{russakovsky2015imagenet}
Russakovsky, O., Deng, J., Su, H., Krause, J., Satheesh, S., Ma, S., Huang, Z.,
  Karpathy, A., Khosla, A., Bernstein, M., et~al.
\newblock Imagenet large scale visual recognition challenge.
\newblock \emph{IJCV}, 2015.

\bibitem[Sablayrolles et~al.(2019)Sablayrolles, Douze, Ollivier, Schmid, and
  J{\'{e}}gou]{sablayrolles19whitebox}
Sablayrolles, A., Douze, M., Ollivier, Y., Schmid, C., and J{\'{e}}gou, H.
\newblock White-box vs black-box: Bayes optimal strategies for membership
  inference.
\newblock In \emph{ICML}, 2019.

\bibitem[Shafahi et~al.(2018)Shafahi, Huang, Najibi, Suciu, Studer, Dumitras,
  and Goldstein]{shafahi18poisonfrogs}
Shafahi, A., Huang, W.~R., Najibi, M., Suciu, O., Studer, C., Dumitras, T., and
  Goldstein, T.
\newblock Poison frogs! targeted clean-label poisoning attacks on neural
  networks.
\newblock In Bengio, S., Wallach, H., Larochelle, H., Grauman, K.,
  Cesa-Bianchi, N., and Garnett, R. (eds.), \emph{NeurIPS}, 2018.

\bibitem[Shokri et~al.(2017)Shokri, Stronati, and
  Shmatikov]{shokri17membershipinference}
Shokri, R., Stronati, M., and Shmatikov, V.
\newblock Membership inference attacks against machine learning models.
\newblock \emph{IEEE Symp. Security and Privacy}, 2017.

\bibitem[Steinhardt et~al.(2017)Steinhardt, Koh, and
  Liang]{steinhardt2017certified}
Steinhardt, J., Koh, P. W.~W., and Liang, P.~S.
\newblock Certified defenses for data poisoning attacks.
\newblock In \emph{NeurIPS}. 2017.

\bibitem[Szegedy et~al.(2014)Szegedy, Zaremba, Sutskever, Bruna, Erhan,
  Goodfellow, and Fergus]{szegedy14intriguing}
Szegedy, C., Zaremba, W., Sutskever, I., Bruna, J., Erhan, D., Goodfellow,
  I.~J., and Fergus, R.
\newblock Intriguing properties of neural networks.
\newblock In \emph{ICLR}, 2014.

\bibitem[Thomee et~al.(2015)Thomee, Shamma, Friedland, Elizalde, Ni, Poland,
  Borth, and Li]{thomee2015yfcc100m}
Thomee, B., Shamma, D.~A., Friedland, G., Elizalde, B., Ni, K., Poland, D.,
  Borth, D., and Li, L.-J.
\newblock Yfcc100m: The new data in multimedia research.
\newblock \emph{arXiv preprint arXiv:1503.01817}, 2015.

\bibitem[Tishby et~al.(2000)Tishby, Pereira, and Bialek]{tishby2000information}
Tishby, N., Pereira, F.~C., and Bialek, W.
\newblock The information bottleneck method.
\newblock \emph{arXiv preprint physics/0004057}, 2000.

\bibitem[Torralba et~al.(2011)Torralba, Efros, et~al.]{torralba2011unbiased}
Torralba, A., Efros, A.~A., et~al.
\newblock Unbiased look at dataset bias.
\newblock In \emph{CVPR}, volume~1, pp.\ ~7, 2011.

\bibitem[Tran et~al.(2018)Tran, Li, and Madry]{tran18spectral}
Tran, B., Li, J., and Madry, A.
\newblock Spectral signatures in backdoor attacks.
\newblock In \emph{NeurIPS}. 2018.

\bibitem[Vukoti{\'c} et~al.(2018)Vukoti{\'c}, Chappelier, and
  Furon]{vukotic2018deep}
Vukoti{\'c}, V., Chappelier, V., and Furon, T.
\newblock Are deep neural networks good for blind image watermarking?
\newblock In \emph{Workshop on Information Forensics and Security (WIFS)}.
  IEEE, 2018.

\bibitem[Yeom et~al.(2018)Yeom, Giacomelli, Fredrikson, and
  Jha]{yeom18privacyrisk}
Yeom, S., Giacomelli, I., Fredrikson, M., and Jha, S.
\newblock Privacy risk in machine learning: Analyzing the connection to
  overfitting.
\newblock In \emph{CSF}, 2018.

\bibitem[Zhu et~al.(2018)Zhu, Kaplan, Johnson, and Fei-Fei]{zhu2018hidden}
Zhu, J., Kaplan, R., Johnson, J., and Fei-Fei, L.
\newblock Hidden: Hiding data with deep networks.
\newblock In \emph{ECCV}, 2018.

\end{thebibliography}

\clearpage

\appendix
\section{Distillation}
\label{app:distillation}

\begin{table}
\centering {\small 
\begin{tabular}{r@{\mysp}r@{\mysp}r@{\mysp}r@{\mysp}r@{\mysp}r}
\toprule
\% radioactive  & 1        &   2      &   5      & 10       &   20   \\
\midrule
$\log_{10}(p)$  &  $-1.58$ & $-3.07$ & $-13.60$  & $-34.22$  & $-137.42$ \\
\bottomrule
\end{tabular}
}
\smallskip
\caption{\label{tab:distillation_results}
$p$-value for the detection of radioactive data usage.
A Resnet-18 is trained on Imagenet from scratch, and a percentage of the training data is radioactive.
This marked network is distilled into another network, on which we test radioactivity.
When $2\%$ of the data or more is radioactive, we are able to detect the use of this data with a strong confidence ($p < 10^{-3}$).
}
\end{table}

Given a marked resnet-18 on which we only have black-box access, we use distillation \cite{hinton2015distilling} to train a second network. 
On this distilled network, we perform the radioactivity test.
We show in Table \ref{tab:distillation_results} the results of this radioactivity test on distilled networks. 
We can see that when $2\%$ or more of the original training data is radioactive, the radioactivity propagates through distillation with statistical significance ($p<10^{-3}$).

\end{document}